\documentclass{article}

\usepackage{arxiv}

\usepackage[utf8]{inputenc} 
\usepackage[T1]{fontenc}    
\usepackage{hyperref}       
\usepackage{url}            
\usepackage{booktabs}       
\usepackage{amsfonts}       
\usepackage{nicefrac}       
\usepackage{microtype}      
\usepackage{lipsum}
\usepackage{algpseudocode}
\usepackage{graphicx}
\usepackage{siunitx}
\usepackage{xcolor}
\usepackage{dcolumn}
\usepackage{array}
\usepackage{multirow}
\usepackage{subcaption}

\title{Headline Generation: Learning from Decomposable Document Titles}

\author{
 Oleg Vasilyev, Tom Grek and John Bohannon \\
  Primer\\
  San Francisco, California\\
  \texttt{\{oleg,tom,john\}@primer.ai} \\
}

\begin{document}
\maketitle
\begin{abstract}
We propose a novel method for generating titles for unstructured text documents. We reframe the problem as a sequential question-answering task. A deep neural network is trained on document-title pairs with decomposable titles, meaning that the vocabulary of the title is a subset of the vocabulary of the document. To train the model we use a corpus of millions of publicly available document-title pairs: news articles and headlines. We present the results of a randomized double-blind trial in which subjects were unaware of which titles were human or machine-generated. When trained on approximately 1.5 million news articles, the model generates headlines that humans judge to be as good or better than the original human-written headlines in the majority of cases.
\end{abstract}

\section{Introduction}
The title of a document can be considered as the shortest possible summary of the document. Automatically generating this concise summary requires extremely low error tolerance. The slightest grammatical deficiency or factual error can render a title functionally useless \cite{lopyrev2015generating, ayana2016neural, kiyono2017source, xu2019novel}. 

The general task of text summarization is well reviewed \cite{khatri2018abstractive, ayana2016neural, allahyari2017text, esmaeilzadeh2019neural} and is divided broadly into two methods: extractive and abstractive. 

Extractive summarization extracts long spans of text from a document, usually whole sentences, and assembles them \cite{erkan2004graph, mihalcea2004textrank}. Although it lacks expressivity, the method has the advantage that sentences are at least guaranteed to be coherent because they were originally written by humans. 

Abstractive summarization does away with this limitation, drawing upon a dictionary \cite{khatri2018abstractive}. In principle this removes limits to expressivity, but it comes at a steep cost. Existing abstractive summarization systems suffer from grammatical errors, factual hallucinations, and incoherence \cite{see2017get, xu2019novel, kiyono2017source, esmaeilzadeh2019neural, lopyrev2015generating}. 

A recent attempt at a compromise between the two methods uses abstractive summarization with an extractive fall-back \cite{see2017get}. By extracting unknown words and phrases from a text with a pointer-generator network, it is possible to avoid some of the errors of abstractive methods.

In this paper we present an approach that completely abandons the use of a dictionary for abstractive text generation. It is generally recognized that this dictionary is a significant source of error \cite{kumar2019}. We instead compose a headline by extracting all necessary words from the text of the document. Even a document of modest length - typical news articles are between 400 and 800 words long - contains sufficient vocabulary to express a functional headline. Our motivation is to force the text generation model to concentrate on the text itself without the distraction of a large external vocabulary.

The method we present thus blurs the distinction between extractive and abstractive summarization. It can be considered either as abstractive with a purpose-limited dictionary provided by the text itself, or as extractive with high flexibility in its choice of text spans and assembly. For training and evaluation of our approach we use Primer’s database of daily news documents.

\section{Our approach}
\subsection{Headline decomposition}
We want select only the documents with  \textit{decomposable} titles, where decomposition is defined by the following greedy longest-match-first algorithm that performs iterative search through the document text:
\begin{enumerate}
\itemsep0em
  \item Find in the text the longest substring of the title; the substring must start at the beginning of the title. If there is more than one substring of of equal length, take the first one encountered from the beginning of the text.
  \item Repeat the following until the end of the title: Slice the title starting from the end of the previously found substring, and repeat step 1 using this slice of the title. If no substring is found in the document text, terminate and discard the title.
\end{enumerate}

More formally: 

The title $H$ is represented as concatenation of strings $H = S_1S_2...S_n$ where each string $S_i$ is as short as possible, providing that there is no any pair $S_i, S_{i+1}$ such that the last character of $S_i$ and the first character of $S_{i+1}$ are both letters or are both digits. 
The text $T$ is represented in the same way: $T =W_1W_2...W_N$.

The pseudo code for the decomposition of the title into text spans is shown in Figure \ref{fig:algorithm}.

 \begin{figure}[h]
     \centering
     \begin{tabular}{l}
     \toprule
     Given a title $S_1S_2...S_n$ and a text $T =W_1W_2...W_N$ \\
     Initialize $Decomposition$ to an empty list \\
     Initialize $q$ to 1 \\
     \bf{while} $q \leq n $:\\
     \hspace{.5cm}$G = S_qS_{q+1}...S_n$ \\
     \hspace{.5cm}Find largest $k\geq q$ to get $D =S_qS_{q+1}...S_k$ such that \\
     \hspace{1.5cm}$G = DS_{k+1}...S_n$  and \\
     \hspace{1.5cm}$T =W_1W_2...W_mDW_{m+k-q+2}...W_N$\\ 
     \hspace{.5cm}If more than one $m$ exists, select the smallest.\\
     \hspace{.5cm}\textbf{if} $k$ not found \textbf{then}\\
     \hspace{.5cm}\hspace{.5cm}$Decomposition$ = empty list\\
     \hspace{.5cm}\hspace{.5cm}break\\
     \hspace{.5cm}\textbf{else}:\\
     \hspace{.5cm}\hspace{.5cm}Add $D$ to $Decomposition$\\
     \hspace{.5cm}\hspace{.5cm}$q = k+1$\\
     \bottomrule
     \end{tabular}
     \caption{For document-title pair, determine if the title is decomposable, and if so provide the decomposition.}
     \label{fig:algorithm}
 \end{figure}

Consider real example of a document with the title "\textit{Christopher John: I am very happy to be here}". The following substrings were sequentially found in the text of the document:

\hspace{0.5cm} "\textit{Christopher John}", "\textit{: }", "\textit{I am very happy to }", "\textit{be here}"

The text did not have the string "\textit{Christopher John:}" nor the string "\textit{I am very happy to be here}". But the text did contain smaller pieces of the title. If any of the smallest pieces did not exist in the text, for example the word ‘happy’ or the ‘:’ punctuation mark, then the document would not have a decomposable title and would not be included in our training data set.

We have observed that about 10\% of titles in the English language media are decomposable.

\subsection{Training set}
We build a training set from the decomposed headlines. The training data are gathered therefore from about 10\% of news documents - such documents that have decomposable titles. For each document, training samples are created as tuples $Sample = (Question, Text, Answer)$, where the $(Question, Text)$ is the input, and the $Answer$ is the output. Gathering the samples from each document is done as described in Figure \ref{fig:trainset}.

 \begin{figure}[h]
     \centering
     \begin{tabular}{l}
     \toprule
     Given a headline decomposition $D_1...D_p$ and a $Text$\\
     Initialize $Samples$ to empty list\\
     Initialize $Question$ to empty string\\
     \textbf{for} $i$ from $1$ to $p$:\\
     \hspace{.5cm}$Answer = D_i$\\
     \hspace{.5cm}Add $(Question, Text, Answer)$ to $Samples$\\
     \hspace{.5cm}Update $Question$ to $Question + Answer$\\
     Add $(Question, Text, \_)$ to $Samples$\\
     \bottomrule
     \end{tabular}
     \caption{Creating training samples from decomposed titles}
     \label{fig:trainset}
 \end{figure}
 
 The last tuple contains the termination symbol '\_'. Thus, each document produces two or more training samples. Two is the minimal possible number: It happens when the whole title is found at the first search, and the termination answer is assigned to the second question. Notice that the very first ‘question’ is always an empty string.
 
 The tuples $(Question, Text, Answer)$ are then used for training a traditional question-answer model. We used the BERT transformer base uncased model \cite{devlin2018bert} adapted to question-answering by the Huggingface team \cite{huggingface}.
 
\subsection{Title Generation}
\begin{figure}[h]
    \centering
    \includegraphics[width=0.6\textwidth]{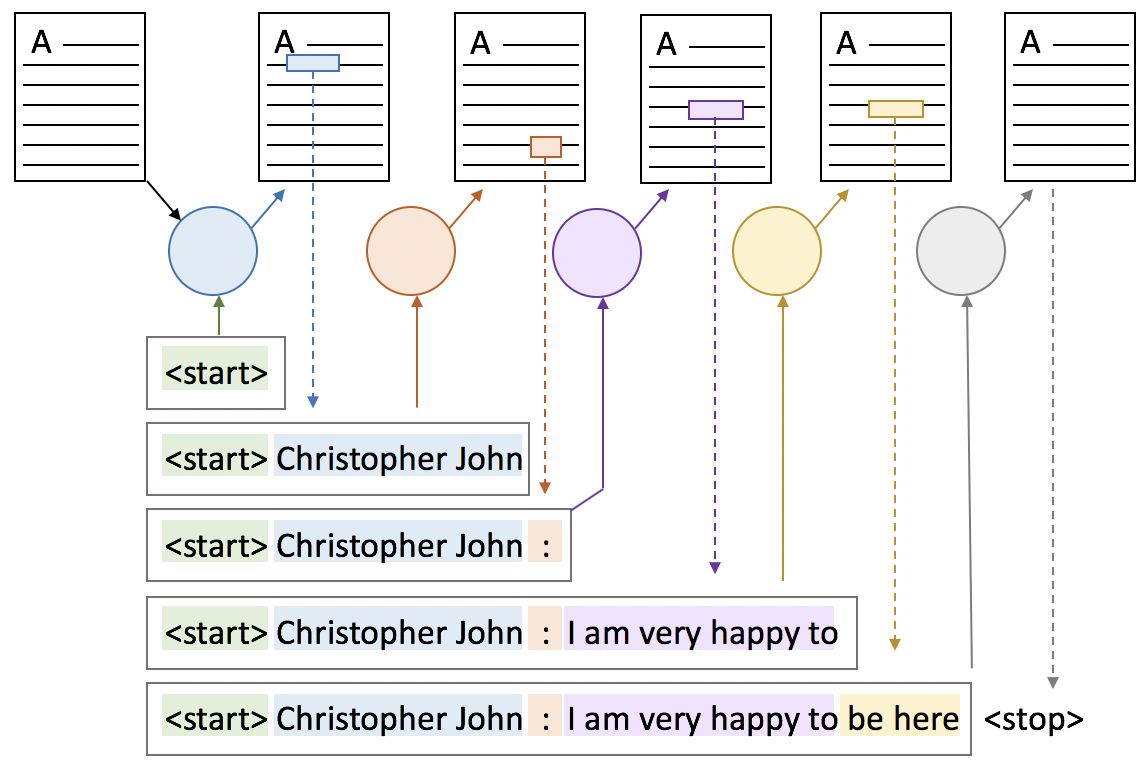}
    \caption{Headline generation as question-answer sequence. The first question is an empty string. The answer is a span in the text. The answer is then concatenated with the question to form the next question. This process is repeated until the answer returned is a termination token.}
    \label{fig:headline_generation}
\end{figure}

At generation time we ask ‘questions’ and accumulate text in the form of ‘answers’ until the termination symbol is an answer. The termination answer, the symbol ‘\_’, is always available because we add it to the end of every document. A title is obtained by concatenating all the answers except the termination.
Returning to our example of a title, Fig \ref{fig:headline_generation} illustrates how the sequence of question-answers looks if such title would be generated from the text.

The model (evaluated as explained in the next section) was trained on a question-answer dataset created from 10 weeks of Primer data, starting October 1, 2018. On a typical day, this data stream consists of between 100K and 200K English-language news documents, of which approximately 10\% enter our training data. The training data set represents 7 million question-answer examples (including the termination answers). The question-answer examples were obtained from 1.5M documents with decomposable titles. The training was done by fine-tuning the BERT transformer uncased base model for 3 epochs. On average, a generation of one headline takes about 0.15 second on GPU V100.  

\begin{figure}[h]
    \centering
    \includegraphics[width=0.85\textwidth]{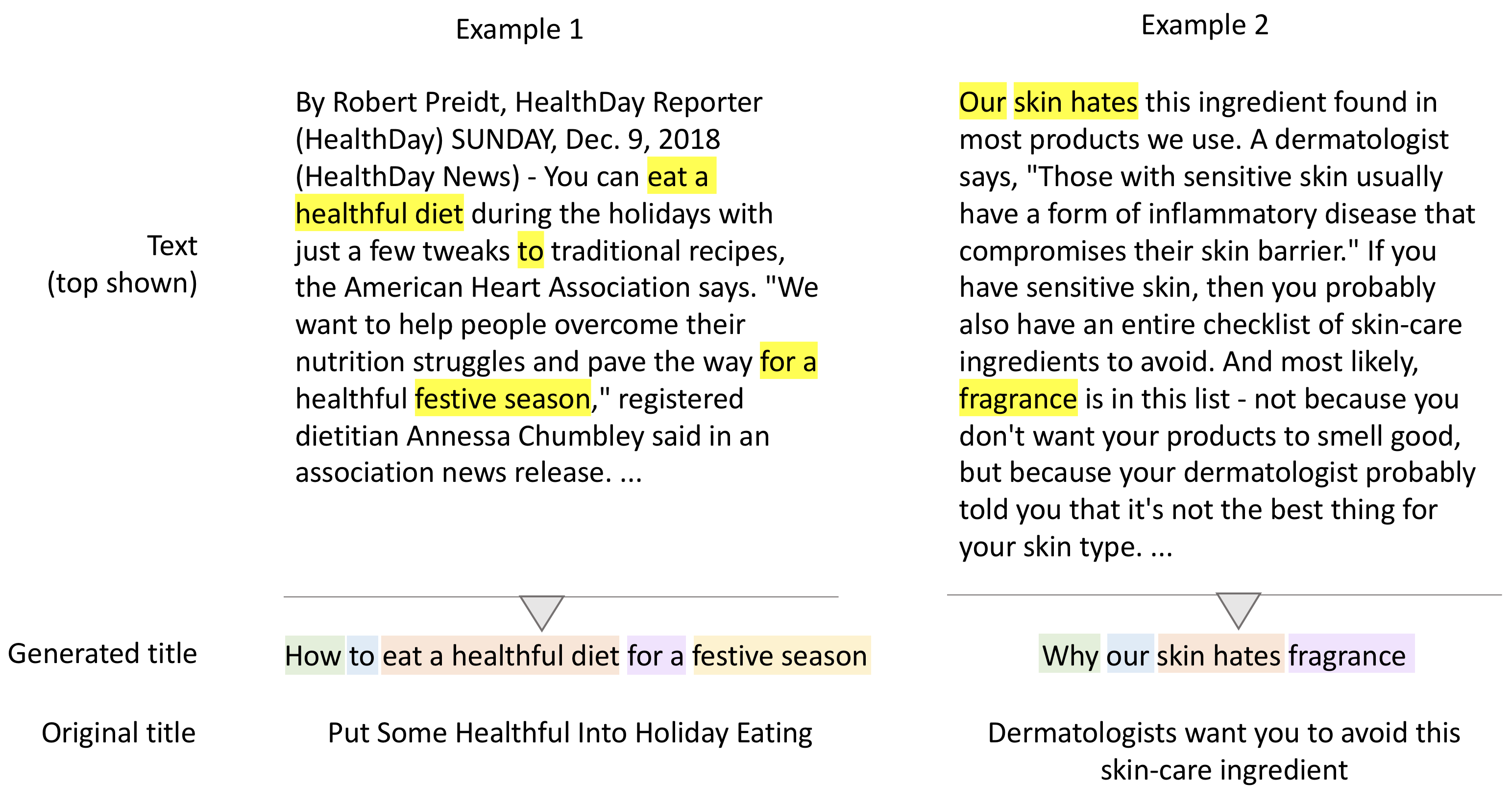}
    \caption{Examples of generated headlines. For each example only the top of the document text used for generation is shown here. The output of each question-answer step in the generated headline is represented by a different highlighting color. All the 'answers' are also highlighted yellow in the text (unless they came from lower parts of text not shown here). For comparison, the original titles are also shown.}
    \label{fig:sample_headlines}
\end{figure}

Examples of real and generated titles are presented in Fig \ref{fig:sample_headlines}.

\textbf{How different are the generated titles from the human-written originals?} We know in advance that about 90\% of generated titles will necessarily differ from the original titles, simply because roughly 90\% of news article titles are not decomposable. But moreover, even for documents with fully decomposable titles, our model generates titles different from the human-written original. It is extremely rare that a generated title is an exact match to the original. On average, a real decomposed title consist of 4 to 5 'answers'. The same is true on average for a generated title, but the 'answers' can come from very different locations in the text.

\textbf{How relevant are specific titles from the latest days of the training data?} We know that in our data less than 0.5\% of daily news titles match titles from the previous week. And those that do overlap are generic titles such as “TV/Radio”, “Tuesday”, and “Tuscarawas County”. Such titles are typically not decomposable. Generally, we find it difficult even to overfit our model on the training set itself. In order to obtain the majority of generated titles repeating the training titles, we had to decrease the training set to thousands of decomposable titles and to increase the number of epochs above ten.

\section{Human evaluation}
Human evaluation is especially important for our headline generation, because we know in advance that our generated titles in most cases should not be close to real titles as measured by automated text overlap metrics such as ROUGE \cite{lin2004rouge}.

\subsection{Evaluation setup}
The evaluation is done at a day after the training time interval ended. For evaluation, we presented to human evaluators the following task. You see the body text of an article and two titles. The task is to independently score the titles, each on their own merit. One of the titles is real while the other is machine-generated (in randomly shuffled order). The titles are scored on a 5-point scale: Very Bad, Bad, OK, Good, Very Good.

The evaluation was done using Prodigy \cite{prodigy}, a spaCy annotation tool. In the absence of codified standards for evaluating models using human graders, we opted to present each text and title pair individually, ensuring that our graders could not examine more than one generated (or real) title simultaneously. We limited the task to 100 documents at a time which takes on average about 1 hour to complete.

The 100 documents for human evaluation were randomly picked as one document per source from popular news sources, on the day after the model training period. The sources used were chosen by ranking those most frequently cited on English Wikipedia, excluding non-journalistic sources and exclusively sports and business-focused sources.

The top five of our selected sources were: (1) New York Times (nytimes.com), (2) Washington Post (washingtonpost.com), (3) BBC (bbc.co.uk), (4) Forbes (forbes.com) and (5) CBC.Ca (cbc.ca).

The evaluation presented here was undertaken by 10 evaluators: 2 authors of this paper and 8 hired external evaluators. Each evaluator scored a total of 200 titles: one real and one generated for each of the 100 documents. Figure \ref{fig:instructions} shows the instructions text for the task.

\begin{figure}[h]
\centering
\fbox{
\begin{minipage}{13 cm}
    You are shown two or more possible headlines for an article. Score each headline for its quality \textbf{independently}. (Some headlines might be \underline{better} than the others, or \underline{worse}, or they could all be the \underline{same quality}.) Sometimes headlines are good, and sometimes they are bad. You must be the judge!

    What makes a \textbf{good} headline? It should be...
    \begin{itemize}
        \item Informative. It should tell you what the article is about, including key details.
        \item Easy to read. It should not be too long or full of extra details.
        \item Well-written. It should not have grammatical errors or awkward wording.
    \end{itemize}
    
    What makes a \textbf{bad} headline?
    \begin{itemize}
        \item Irrelevant details included.
        \item \textcolor{red}{Factually incorrect}. (This is the worst of all!)
    \end{itemize}
        \end{minipage}
}
    \caption{Instructions for evaluating headlines.}
    \label{fig:instructions}
\end{figure}

\subsection{Score distribution}
\begin{figure}[h]
    \begin{tabular}{p{.575\textwidth} p{.001\textwidth}p{.425\textwidth}}
    \vspace{0pt} \includegraphics[width=0.575\textwidth]{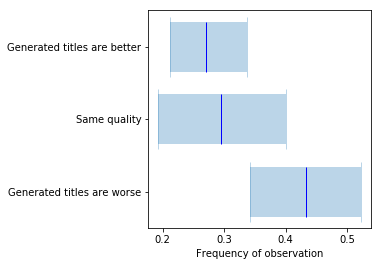}
    &   &
    \vspace{0pt} \includegraphics[width=0.425\textwidth]{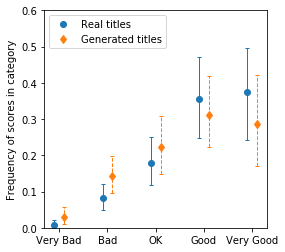} \\
    (a) Normalized distribution of the comparison between generated and real titles, with 95\% confidence intervals. The scores are given by the same person to the titles for the same texts. & &
    (b) Normalized distribution of the scores given by human evaluators to generated titles (dotted orange, with diamond marker) and real titles (continuous blue, with circle marker), with 95\% confidence intervals shown.
    \end{tabular}
    \caption{Analysis of bootstrap results}
    \label{fig:bootstrap_results}
\end{figure}

In order to produce an estimate from the distribution of the gathered scores, we performed Bootstrap with 1 million samples, where each sampling was obtained by two mutually independent random selections with replacement: selection of the evaluators and selection of the documents. The results of comparison of scores given by the same evaluator to the real vs. generated titles of the same document are shown in Figure \ref{fig:bootstrap_results}(a), with 95\% confidence intervals. The distribution of the scores with 95\% confidence interval is shown in Figure \ref{fig:bootstrap_results}(b).

\begin{figure}
    \centering
    \includegraphics[width=10cm]{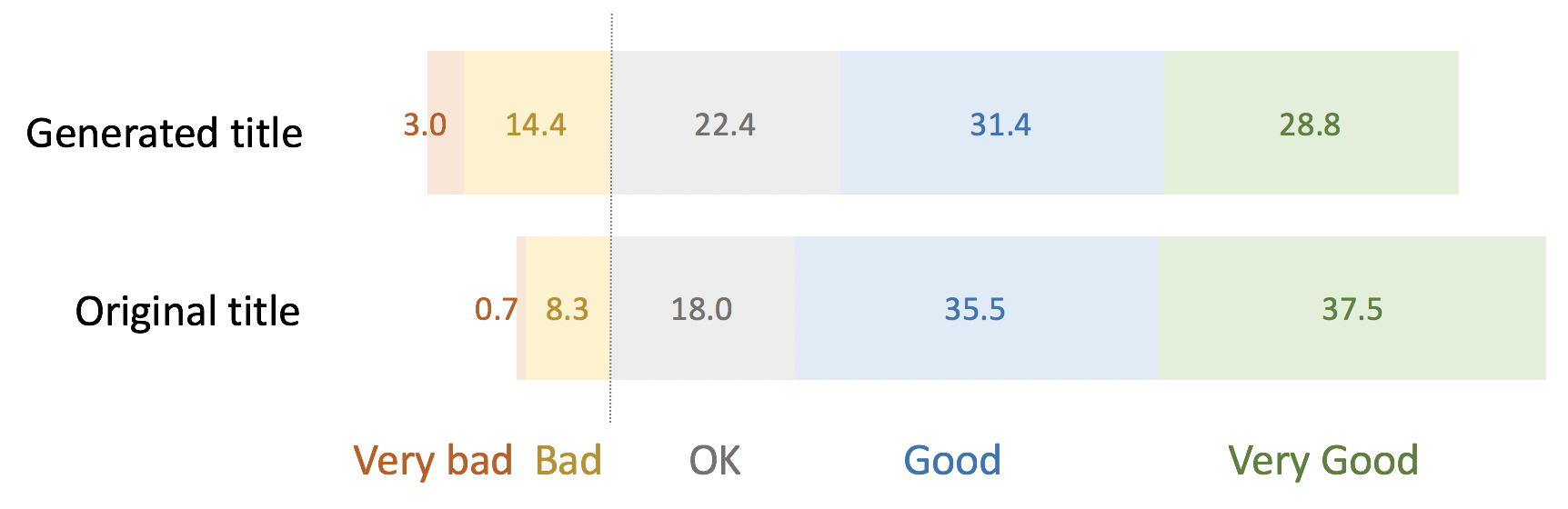}
    \caption{Evaluation of the titles on a 5 tier scale. Median values obtained from the Bootstrap distribution are shown for all 5 possible outcomes (ordered visually from the worst score - "very bad" in red - to the best - "very good" in green).}
    \label{fig:median_value}
\end{figure}

Figure \ref{fig:median_value} is equivalent to the Figure \ref{fig:bootstrap_results}(b) stripped of the confidence intervals and puts the visual emphasis on comparing the quality of the titles.

\begin{samepage}
A couple more characterizations from the Bootstrap are of practical interest: The real title is $OK$ or better, while the generated title is $Bad$ or worse in 17\% of cases. The generated title is $OK$ or better, while the real title is $Bad$ or worse in 9\% of cases.
\end{samepage}

The Table \ref{tab:titles_scores} shows several examples of differently scored real and generated titles, in order to illustrate the evaluation scores.
\begin{table}
 \caption{Examples of real and generated titles, with median scores. The shown scores are medians, taken from the scores of all evaluators with 0 the worst score and 4 the best. For instance, the underlying scores for the first row is (0, 1, 1, 1, 1, 1, 2, 2, 2, 2) in the case of a real title and (2, 2, 2, 3, 3, 3, 3, 3, 3, 4) for the generated title.}
  \centering
  \begin{tabular}{p{7.8cm}p{8cm}}
    \toprule
    \begin{tabular}{p{1.2cm}p{4.8cm}p{1cm}}
        &
        \bf Title & 
        \bf Median score 
    \end{tabular}
    & 
    \begin{tabular}{p{6.7cm}}
        \bf Text: only the top is shown here \\ \\
    \end{tabular}\\
    \toprule
    \begin{tabular}{r p{5cm} c}
        real & Postal pension & 1.0 \vspace{4pt} \\
        \midrule
        generated & No acknowledgement of pension amendments & 3.0
    \end{tabular}
    &
    \begin{tabular}{p{7.9cm}}
    Dear Claudienne I worked with the Post and Telecommunications Department in St Catherine from 1968 to 1977. When I applied for my pension the Post Office headquarters sent my file to the Ministry of Finance (MOF) on October 5, 2016. However, to date (October 13, 2017) I have received no letter of acknowledgement ...
    \end{tabular} \\
    \bottomrule
    \begin{tabular}{r p{5cm} c}
        real & It's been a long 2 years' for Kelly, Kudlow says & 2.0 \vspace{4pt} \\
        \midrule
        generated & John Kelly replacement to be announced in next few days & 3.5
    \end{tabular}
    &
    \begin{tabular}{p{7.9cm}}
    White House economic adviser Larry Kudlow, right, said the replacement for chief of staff John Kelly, center, would be announced Monday or early in the week. | Mark Wilson/ Share on Facebook Share on Twitter It is unclear if White House chief of staff John Kelly decided to resign from his post or was forced to leave ...
    \end{tabular} \\
    \bottomrule
    \begin{tabular}{r p{5cm} c}
        real & Bad Movie Diaries: A Christmas Prince: The Royal Wedding (2018) & 2.5 \vspace{4pt} \\
        \midrule
        generated & Jim Vorel and Kenneth Lowe discuss A Christmas Prince and its sequel, The Royal Wedding & 4.0
    \end{tabular}
    &
    \begin{tabular}{p{7.9cm}}
Jim Vorel and Kenneth Lowe are connoisseurs of terrible movies. In this occasional series , they watch and then discuss the fallout of a particularly painful film. Be wary of spoilers. Ken: A very happy holiday to you again, Jim. As I sip my eggnog here in the winter wonderland that is 50-degree, tornado-ravaged Central Illinois ...
    \end{tabular} \\
    \bottomrule
    \begin{tabular}{r p{5cm} c}
        real & Moxie Girl cleaning company experiences growth amidst vacation rental market boom & 4.0 \vspace{4pt}\\
        \midrule
        generated & Moxie Girl is turning around homes quickly & 2.0
    \end{tabular}
    &
    \begin{tabular}{p{7.9cm}}
Unless it\'s messy, dirty or somewhat disheveled, vacation-home renters probably don\'t notice when anything\'s amiss at their home away from home and instantly feel, well, at home. Amanda Thomas, the CEO and founder of Moxie Girl, makes sure that happens. The influx of investment properties used for vacation ...
    \end{tabular} \\
    \bottomrule
    \begin{tabular}{r p{5cm} c}
        real & Schiff Says Trump May 'Face The Very Real Prospect Of Jail Time' & 4.0 \vspace{4pt}\\
        \midrule
        generated & Schiff: Trump will have jail time & 1.5
    \end{tabular}
    &
    \begin{tabular}{p{7.9cm}}
The congressman expected to become the new chairman of the House Intelligence Committee is predicting dark days ahead for President Donald Trump, including potential jail time. Democratic Rep. Adam Schiff of California was on "Face the Nation" ...
    \end{tabular} \\
    \bottomrule
    \begin{tabular}{r p{5cm} c}
        real & Motorist shot on West Capitol Drive, street closed in both directions & 4.0 \vspace{4pt}\\
        \midrule
        generated & Motorist shot in traffic accident on West Capitol Drive, shut both directions after motorist shot & 1.0
    \end{tabular}
    &
    \begin{tabular}{p{7.9cm}}
Share This Story! Let friends in your social network know what you are reading about A motorist was shot on W. Capitol Dr. near N. 7th St. Sunday morning. Post to Facebook Sent! A link has been sent to your friend's email address. Posted! A link has been posted to your Facebook feed. A traffic accident on West Capitol ...
    \end{tabular} \\
    \bottomrule
  \end{tabular}
  \label{tab:titles_scores}
\end{table}  

Assessment of headline quality is highly subjective, so the inter-rater reliability is low. The Krippendorff's alpha \cite{krippendorff2018content, fastkrippendorff} for scores considered as intervals is 0.27. If the scores for generated and real headlines are considered separately, the alpha is higher for the generated headlines 0.31, and lower for the real headlines 0.17. This may reflect the fact that the generated headlines closer follow the text (by the nature of our generation algorithm) and this makes the evaluator judgment more certain.

The real vs generated comparison rating (with values -1,0,1 as worse, same or better) has also low inter-rater reliability: the alpha is 0.23. The low inter-rater reliability reflects low agreement between the human evaluators but does not negate the above results representation obtained with Bootstrap, since the samplings do include selection of evaluators.

\section{Do we miss having a dictionary?}

\subsection{Expanding the definition of decomposability}

In the previous section, we obtained sequences of question-answer from a document - headline pair only when the vocabulary of the document headline was a subset of the vocabulary of the document body. This means in particular that if at least one word was not found in the text, nothing from that title would enter our training set. 

Let us consider now a possibility of expanding the definition by adding a dictionary, and allowing the title decomposition to draw a word from the dictionary if the word is not found in the text. At generation time, we assume that the dictionary could be used in a manner similar to a pointer-generator network \cite{see2017get}, so that a word from the dictionary could be picked instead of a span from the text when this provides a better generation path.

The motivation to adding a dictionary is to increase flexibility of the text generation. However, we hypothesize that access to a dictionary would damage the model's performance by causing text generation errors \cite{kumar2019}. Note that although this relaxation of decomposabiity will lead to an increase of the fraction of decomposable titles, the possible benefit here is not in increasing the training set (which can be made as large as needed either way) but in increasing the variety of the training samples.   

\subsection{Defining a dictionary}

We focus on defining a vocabulary of words that typically appear in titles but not in the corresponding document body. We then perform an additional filtering step to limit the risk of having our algorithm hallucinate statements which are not supported by facts in the text.

We restrict ourselves to word tokens consisting of alphabetic characters. We consider 6 months of news articles (Jun - Nov 2018). There are 16.2 million distinct words in the texts of the articles. This 'in-text' dictionary of course has a familiar distribution of words, with ‘the’, ‘of’, ‘and’ at the top. There are 2.2 million distinct words in the titles of the articles. This ‘in-title’ dictionary is similar, except that the names of month come close to the top. Finally, there are 0.8 million distinct words which are in the titles but not in the corresponding texts. This ‘in-title-not-in-text’ dictionary is profoundly different. Names of the months and some special time-related abbreviations rise to the top, for example ‘June’, ‘EDT’, ‘CEST’ etc. We argue that it is better to pick such words from the text rather than to ‘hallucinate’ them from an external dictionary. Hence our strategy for creating and filtering of the dictionary described in Figure \ref{fig:create_dictionary}.

\begin{figure}[h]
    \centering
    \begin{tabular}{c p{12cm}}
        \toprule
        $1.$ & Create a dictionary (with counts) of cased words found in title but not in text.  \\
        $2.$  & If a word has a count of at least 100 occurrences as all lowercase (in title) then combine the uppercase and lowercase counts and keep the word as lowercase.\\
        $3.$  & Remove all words that are not lowercase.\\
        $4.$  & Sort by counts of occurrences.\\
        $5.$  & Select top $N$.\\
        \bottomrule
    \end{tabular}
    \caption{Steps to create and filter the dictionary used for expanding the headline decomposition.}
    \label{fig:create_dictionary}
\end{figure}

After the filtering step, our proposed dictionary starts with following 10 top words (all with counts higher than 200 000):

\begin{center}
\begin{tabular}{l l}
     1. seeks & 6. issues \\
     2. head & 7. decreases \\
     3. increases &  8. says  \\
     4. us & 9. publishes  \\
     5. line & 10. announces \\
\end{tabular}
\end{center}

\subsection{Preliminary results}
\begin{figure}[h]
    \begin{tabular}{p{0.5\textwidth} p{0.5\textwidth}}
         \vspace{0pt} \includegraphics[width=0.45\textwidth]{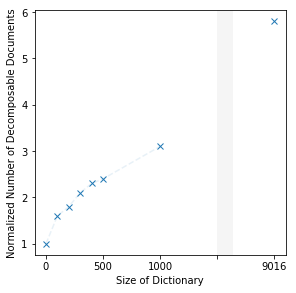}&
         \vspace{0pt} \includegraphics[width=0.45\textwidth]{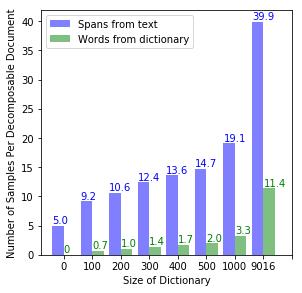}\\
         & 
    \end{tabular}
    \centering
    \caption{(a) Number of documents with decomposable headlines as the dictionary size increases, normalized by the number of documents with decomposable headlines without a dictionary.  (b) Number of training samples per document with decomposable title: coming from text spans is blue; coming from the dictionary is green. }
    \label{fig:dictgrowth}
\end{figure}

 How large would an external dictionary have to be to be useful for the task of headline generation?  
If we keep all the words above the count 500 from our 6-months sampling, the dictionary size is 9016. The bigger the dictionary, the more titles become decomposable not purely by spans from the text but also by including the dictionary words. Figure \ref{fig:dictgrowth} shows how increase of the dictionary size affects our training set.

With adding only the top 100 words of the dictionary, the number of decomposable documents jumps by 60\%, mostly due to ‘saving’ a decomposition of a title by some single word from the dictionary. Most of a title is still composed of spans, with less than 8\% of samples obtained from the dictionary. Notice that the number of samples has grown more than the number of decomposable titles. This means that the added titles have a more complex decomposition strategy and this may yield a steep cost both for training and generation.

With further addition of more words, the training set changes far less dramatically. In principle, an increase of the dictionary size should improve flexibility of title generation, but also increase the probability of hallucinations and incoherence. From the growth of the training patterns we observe, it is reasonable to assume that the added dictionary should be kept small, limited to the top 1000 words or even to the first 100 - 200.

We have not fully explored this direction, but our attempt to add a dictionary of 500 words caused such deterioration of quality of generated titles (with the same amount of training data) that it did not merit an evaluation. The model we used for that occasion used a simple combination of the question-answer output and the dictionary output (the latter as a softmax classifier) on top of BERT.

Our intuition is that we do not need a dictionary for this task. The vocabulary needed for expression of a functional document title can be found in the text itself.

\section{Conclusion}
In this paper we demonstrated a viable approach for generating document headlines. We specifically targeted for a method that would produce factually correct and informative headlines. To this end, we arranged a question-answer model to learn from decomposable titles, and then to generate the headline by answers, without using a dictionary. 

We found from human evaluation of our method that the original headlines are still scored higher in quality than generated ones on average. This is true regardless of the fact that it is difficult to guess whether a headline is real or generated. Curiously, we observe that our generation produces rather unique, yet sensible, headlines. They are distinct from real titles. Even when generated for the documents used in the training set, the generated titles are usually distinct from the real ones.

Further experimenting has also shown us that our model, despite being trained only on news articles, generates functional titles when applied to other document types such as emails, movie plot summaries, and even legal documents. For example, when given the full LaTex source of this paper, the model is not confused by the LaTeX commands (which it has never seen) and produces the title of this paper verbatim. The model of course considers the title string as a very informative statement positioned below far less informative specifications of user packages, and above the author, abstract, and introduction sections. If everything above the abstract is removed, the model generates the title "Generating titles for unstructured text documents", and if the abstract is also removed, the model generates the title "Text summarization" (for version 2 it generated, more creatively, "Extractive summarization from text during headline generation").

We are also exploring (with encouraging results) the use of this headline generation model for generating a bullet-point summary of text, generating headlines for appropriately identifying segments of the text, - this is a work in progress. Finally, we expect that simply using larger models and larger and better filtered data may deliver still better results.

\section{Acknowledgements}
We thank Anna Venancio-Marques for comprehensive review and so much improvement, including figures and tables, and Ethan Chan and Delenn Chin for many helpful comments. And most of all we thank the journalists of the world who created the data and continue to inform the public.

\bibliographystyle{unsrt}

\end{document}